\documentclass[fleqn]{article}

\usepackage{microtype}
\usepackage{graphicx}
\usepackage{subcaption}
\usepackage{booktabs} 

\usepackage{amsmath,amsfonts,bm}

\def\figref#1{figure~\ref{#1}}

\def\secref#1{section~\ref{#1}}

\def\supref#1{supplementary section~\ref{#1}}
\def\eqref#1{equation~\ref{#1}}

\def\1{\bm{1}}

\DeclareMathAlphabet{\mathsfit}{\encodingdefault}{\sfdefault}{m}{sl}
\SetMathAlphabet{\mathsfit}{bold}{\encodingdefault}{\sfdefault}{bx}{n}

\newcommand{\KL}{D_{\mathrm{KL}}}

\newcommand{\br}[1]{\left( {#1} \right)}
\renewcommand{\KL}[2]{\textrm{KL}\!\br{#1\vert\vert#2}}
\newcommand{\JS}[2]{\textrm{JS}\!\br{#1\vert\vert#2}}
\newcommand{\const}{{const.}}
\newcommand{\beq}{\begin{equation}}

\newcommand{\eeq}{\end{equation}}

\newcommand{\f}{\mathnormal{f}}

\renewcommand{\v}[1]{#1}
\newcommand{\cb}[1]{\left\{ {#1} \right\}}

\newcommand{\fdiv}[2]{\textrm{D}_f\!\br{#1\vert\vert#2}}
\newcommand{\sfdiv}[2]{\textrm{D}'_f\!\br{#1\vert\vert#2}}

\newcommand{\tabref}[1]{table \ref{#1}}
\renewcommand{\eqref}[1]{(\ref{#1})}

\newcommand{\ndist}[3]{{\cal{N}}\br{#1\thinspace\vline\thinspace #2,#3}}

\usepackage{hyperref}
\usepackage{url}
\usepackage{graphicx}
\usepackage[utf8]{inputenc} 
\usepackage[T1]{fontenc}    
\usepackage{hyperref}       
\usepackage{url}            
\usepackage{booktabs}       
\usepackage{amsfonts}       
\usepackage{nicefrac}       
\usepackage{microtype}      
\usepackage{wrapfig}
\usepackage{graphicx,url}
\usepackage{amsmath, amssymb}
\usepackage{mathtools}
\usepackage{verbatim}
\usepackage{bigints}
\usepackage[capbesideposition=outside,capbesidesep=quad]{floatrow}
\usepackage{sidecap}
\usepackage{float}

\usepackage{tikz,wrapfig}
\usetikzlibrary{arrows,shapes,backgrounds,through,shadows}
\tikzstyle{cont}=[circle, draw,thick,minimum size=7.5mm,line width=1pt,>=stealth]  
\tikzstyle{obs}=[fill=blue!10,thick]  
\tikzstyle{contobs}+=[cont]
\tikzstyle{contobs}+=[obs]
\tikzstyle{discobs}+=[disc]
\tikzstyle{discobs}+=[obs]
\tikzstyle{dgraph}=[->, line width=1.5pt]

\usepackage{hyperref}

\newcommand{\KLs}[2]{\textrm{KL}'\!\br{#1\vert\vert#2}}
\renewcommand{\KL}[2]{\textrm{KL}\!\br{#1\vert\vert#2}}

\renewcommand{\v}[1]{#1}
\renewcommand{\eqref}[1]{(\ref{#1})}

\usepackage[accepted]{icml2019}

\icmltitlerunning{Variational $\f$-divergence Minimization}

\begin{document}
\twocolumn[
\icmltitle{Variational $\f$-divergence Minimization}




\begin{icmlauthorlist}
\icmlauthor{Mingtian Zhang}{to}
\icmlauthor{Thomas Bird}{to}
\icmlauthor{Raza Habib}{to}
\icmlauthor{Tianlin Xu}{goo}
\icmlauthor{David Barber}{to}
\end{icmlauthorlist}

\icmlaffiliation{to}{University College London, United Kingdom}
\icmlaffiliation{goo}{ London School of Economics, United Kingdom}

\icmlcorrespondingauthor{Mingtian Zhang}{mingtian.zhang.17@ucl.ac.uk}

\icmlkeywords{Machine Learning, ICML}

\vskip 0.3in
]



\printAffiliationsAndNotice{} 
\begin{abstract}

Probabilistic models are often trained by maximum likelihood, which corresponds to minimizing a specific $\f$-divergence between the model and data distribution. In light of recent successes in training Generative Adversarial Networks, alternative non-likelihood training criteria have been proposed. Whilst not necessarily statistically efficient, these alternatives may better match user requirements such as sharp image generation. A general variational method for training probabilistic latent variable models using maximum likelihood is well established; however, how to train latent variable models using other $f$-divergences is comparatively unknown. We discuss a variational approach that, when combined with the recently introduced Spread Divergence, can be applied to train a large class of latent variable models using any $f$-divergence.


\end{abstract}
\section{Introduction}
\label{sec:fdiv}

Probabilistic modelling generally deals with the task of trying to fit a model $p_\theta(\v{x})$ parameterized by $\theta$ to a given distribution $p(\v{x})$. To fit the model we often wish to minimize some measure of difference between $p_\theta(\v{x})$ and $p(\v{x})$. A popular choice is the class of $\f$-divergences\footnote{The definition extends naturally to distributions on discrete $x$.} (see for example \cite{DBLP:journals/corr/SasonV15}) which, for two distributions $p(x)$ and $q(x)$, is defined by
\begin{equation}
    \fdiv{p(x)}{q(x)} = \int  q(\v{x})f\br{\frac{p(\v{x})}{q(\v{x})}} d \v{x}
\end{equation}
where $f(x)$ is a convex function with $f(1) = 0$.


Many of the standard divergences correspond to simple choices of the function $\f$, see \tabref{table:div}.
The divergence $\fdiv{p(x)}{q(x)}$ is zero if and only if $p(x)= q(x)$. However, for a constrained model $p_\theta(x)$ fitted to a distribution $p(x)$ by minimizing $\fdiv{p_\theta(x)}{p(x)}$, the resulting optimal $\theta$ can be heavily dependent on the choice of the divergence function $\f$ \citep{minka}.

\iftrue
\begin{table*}[t]
\scalebox{0.85}{
\begin{tabular}{ l | l | l }
  Name & $\fdiv{p(x)}{q(x)}$ & $f(u)$ \\\hline
  forward KL& $\int p(x)\log\frac{p(x)}{q(x)}dx$ & $u\log u$ \\
  reverse KL & $\int q(x)\log\frac{q(x)}{p(x)}dx$ & $-\log u$ \\
  Jensen-Shannon & $\int\cb{\frac{1}{2}p(x)\log\frac{2p(x)}{p(x)+q(x)}+q(x)\log\frac{2q(x)}{p(x)+q(x)}} dx$ & $-(u+1)\log\frac{1+u}{2}+u\log u$\\
  GAN & $\int\cb{ p(x)\log\frac{2p(x)}{p(x)+q(x)}+q(x)\log\frac{2q(x)}{p(x)+q(x)}} dx$ & $-(u+1)\log(1+u)+u\log u$
\end{tabular}}
\caption{Some standard $\f$-divergences. Here $p(x)$ is given and $q(x)$ is the model.  From \citet{nowozin2016f}.\label{table:div}}
\end{table*}
\fi

Whilst there is significant recent interest in using $\f$-divergences to train complex probabilistic models \citep{nowozin2016f}, the $\f$-divergence is generally computationally intractable for such complex models. We consider an upper bound on the $\f$-divergence and show how this bound can be readily applied to training generative models.


\subsection{Maximum likelihood and forward KL}
For data $x_1,\ldots, x_N$ drawn independently and identically from some empirical distribution $\hat{p}(x)$, 
\begin{equation*}
\hat{p}(x) \equiv  \frac{1}{N}\sum_{n=1}^N\delta\br{x-x_n}
\end{equation*}
the forward KL between  an approximating distribution $p_\theta(x)$  and $\hat{p}(x)$ is
\begin{equation}
\KL{\hat{p}(x)}{p_\theta(x)} = -\sum_{n=1}^N \log p_\theta(x_n)+\const
\end{equation}
Minimizing $\KL{\hat{p}(x)}{p_\theta(x)}$ w.r.t. $\theta$ is therefore equivalent to maximizing the likelihood of the data. Given the asymptotic guarantees of the efficiency of maximum likelihood \citep{wolfowitz}, the forward KL is a standard divergence used in statistics and machine learning.

For latent variable models $p_\theta(x) = \int p_\theta(x|z)p(z)dz$, the likelihood objective is usually intractable and it is standard to use the variational evidence lower bound (ELBO)
%
\begin{align}
\begin{split}
\sum_{n=1}^N\log p_\theta(x_n) &\geq \sum_{n=1}^N\int q_\phi(z_n|x_n)[\log{p_\theta(x_n|z_n)p(z_n)}\\&-\log q_\phi(z_n|x_n)]dz_n \equiv L(\theta,\phi)
\end{split}
\label{eq:elbo}
\end{align}
where the so-called variational distribution $q_\phi(z|x)$ is chosen such that the bound (and its gradient) is either computationally tractable or can be readily estimated by sampling \citep{vae}. The parameters $\phi$ of the variational distribution $q_\phi(z|x)$ and parameters $\theta$ of the model $p_\theta(x)$ are jointly optimized to increase the log-likelihood lower bound $L(\theta,\phi)$. 
This lower bound on the likelihood corresponds to an upper bound on the forward divergence $\KL{\hat{p}(x)}{p_\theta(x)}$.

\subsection{Forward versus reverse KL}
It is interesting to compare models trained by the forward and reverse KL divergences. For example, when $p_\theta$ is Gaussian with parameters $\theta=\br{\mu,\sigma^2}$, then minimizing the forward KL gives
\begin{align}
& \arg\min_{\mu,\sigma^2}\KL{p(x)}{p_\theta(x)}\\\Rightarrow \hspace{2mm}& \mu = \int p(x) x dx, \hspace{2mm} \sigma^2 = \int p(x) (x-\mu)^2 dx
\end{align}
%
%

\begin{figure}[ht]
\begin{center}
\centerline{\includegraphics[width=0.5\textwidth]{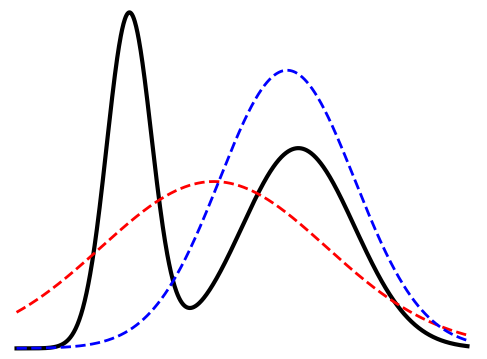}}
\vskip -0.2in
\caption{Fitting a Gaussian to a mixture of Gaussians (black) by minimizing the forward KL (red) and the reverse KL (blue).}
\label{fig:fit:mog}
\end{center}
\vskip -0.2in
\end{figure}
so that the optimal setting is for $\mu$ to be the mean of $p(x)$ and $\sigma^2$ the variance. For an "under-powered" model $p_\theta(x)$ (a model which is not rich enough to have a small divergence) and multi-modal $p(x)$ this could result in $p_\theta(x)$ placing significant mass on low probability regions in $p(x)$. This is the so-called "mean matching" behavior of $\KL{p(x)}{p_\theta(x)}$ that has been suggested as a possible explanation for the poor fidelity of images generated by models $p_\theta(x)$ trained by forward KL minimization \citep{goodfellow2016nips}.
Conversely, when using the reverse KL objective, $\KL{p_\theta(x)}{p(x)}$, for a Gaussian $p_\theta(x)$ and multi-modal $p(x)$ with well separated modes, optimally $\mu$ and $\sigma^2$ fit one of the local modes. This behavior is illustrated in \figref{fig:fit:mog} and is the so-called "mode matching" behavior of $\KL{p_\theta(x)}{p(x)}$. For this reason, the reverse KL  objective has been suggested to be more useful than forward KL divergence if high quality samples are preferable to coverage of the dataset.

This highlights the potentially significant difference in the resulting model that is fitted to the data, depending on the choice of divergence \citep{minka}. In this sense, it is of interest to explore fitting generative models $p_\theta(x)$ to a data distribution $\hat{p}(x)$ using $\f$-divergences other than the forward KL divergence (maximum likelihood).

\subsection{Optimizing $f$-divergences}\label{sec:estmation_f}
Whilst the above upper bound \eqref{eq:elbo} on the forward divergence $\KL{\hat{p}(x)}{p_\theta(x)}$ is well known, an upper bound on other $\f$-divergences (e.g. reverse KL divergence) seems to be unfamiliar \citep{DBLP:journals/corr/SasonV15} and we are unaware of any upper bound on general $\f$-divergences that has been used within the machine learning community.

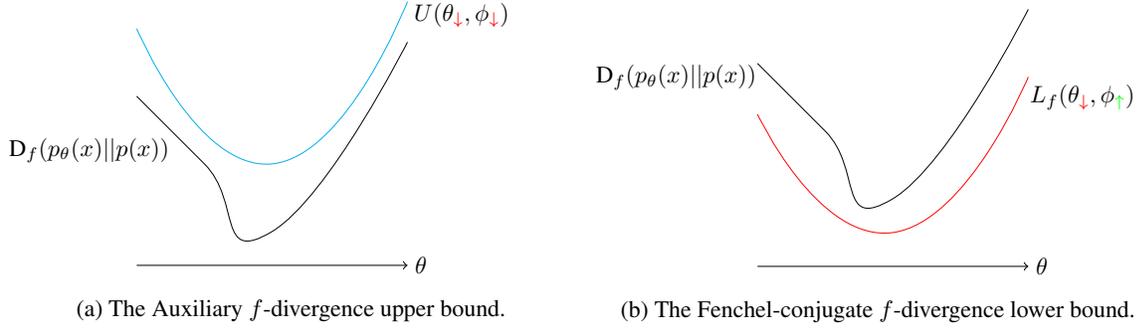
\begin{figure*}[t]
\begin{subfigure}[0]{0.45\textwidth}
\scalebox{0.9}{\begin{tikzpicture}
\draw [] plot [smooth, tension=1] coordinates { (0,0) (1,-1) (2,-2) (4,0.8)};
\node[]() at (-0.7,-0.8) {$\fdiv{p_\theta(x)}{p(x)}$};
\draw [cyan] plot [smooth, tension=1] coordinates { (0,1) (2,-1.) (4,1.4)};
\node[]() at (4.8,1.2) {$U(\theta_{\color{red}{\downarrow}},\phi_{\color{red}{\downarrow}})$};
\draw[->](0,-2.5)--(4,-2.5); \node[]() at (4.2,-2.5) {$\theta$};
\end{tikzpicture}}
\caption{The Auxiliary $\f$-divergence upper bound.}
\end{subfigure}
\begin{subfigure}[1]{0.45\textwidth}
\vspace{0.2cm}
\scalebox{0.9}{\begin{tikzpicture}
\draw [] plot [smooth, tension=1] coordinates { (0,0) (1,-1) (2,-2) (4,0.8)};
\node[]() at (-1.2,-0.2) {$\fdiv{p_\theta(x)}{p(x)}$};
\draw [red] plot [smooth, tension=1] coordinates { (0,-0.75) (2,-2.5) (4,-0.2)};
\node[]() at (4.8,-0.5) {$L_f(\theta_{\color{red}{\downarrow}},\phi_{\color{green}{\uparrow}})$};
\draw[->](0,-3)--(4,-3); \node[]() at (4.2,-3) {$\theta$};
\end{tikzpicture}}
\caption{The Fenchel-conjugate $\f$-divergence lower bound.}
\end{subfigure}
\caption{Upper and lower bounds on the divergence $\fdiv{p_\theta(x)}{p(x)}$. In our upper bound, both the model parameters $\theta$ and bound parameters $\phi$ are adjusted to push down the upper bound, thereby driving down the divergence. In the Fenchel-conjugate approach \cite{nowozin2016f}, the lower bound is made tighter adjusting the bound parameters $\phi$ to push up the bound towards the true divergence, whilst then minimizing this with respect the model parameters $\theta$.
\label{fig:bounds}}
\end{figure*}

Recently a \emph{lower bound} on the $\f$-divergence was introduced in \cite{nowozin2016f,nguyen2010estimating} by the use of the Fenchel conjugate. The resulting training algorithm is a form of minimax in which the parameters $\phi$ that tighten the bound are adjusted so as to push up the bound towards the true divergence, whilst the model parameters $\theta$ are adjusted to lower the bound. \citet{nowozin2016f} were then able to relate the Generative Adversarial Network (GAN) \citep{goodfellow2016nips} training algorithm to the Fenchel conjugate lower bound on a corresponding $\f$-divergence, see \tabref{table:div}.
However, if the interest is purely on minimizing an $\f$-divergence, it is arguably preferable to have an \emph{upper bound} on the divergence since then standard optimization methods can be applied, resulting in a stable optimization procedure, see \figref{fig:bounds}. 


\section{The $\f$-divergence upper bound\label{sec:bound}}


Following the data processing inequality (see for example \cite{renyi}), we obtain an upper bound
%
\begin{equation}
    \fdiv{p(\v{x})}{q(\v{x})} \leq \fdiv{p(\v{x}, \v{z})}{q(\v{x}, \v{z})} \label{eq:f_bound}
\end{equation}
where $p(x,z)$ is a distribution with marginal $\int p(x,z)dz=p(x)$ and similarly $\int q(x,z)dz=q(x)$.
The bound corresponds to a generalization of the auxiliary variational method \citep{auxvar} and can be readily verified using Jensen's inequality:
\begin{align*}
    &\fdiv{p(x, z)}{q(x, z)} \\
    &= \int q(x) \int q(z|x) f\br{\frac{p(x, z)}{q(x, z)}} dz dx \\
    &\geq  \int q(x)  f\br{ \int q(z|x) \frac{p(x, z)}{q(z|x)q(x)} dz} dx \\
    &= \int  q(x) f\br{\frac{p(x)}{q(x)}} dx = \fdiv{p(x)}{q(x)}
\end{align*}
%
Additional properties of the auxiliary $\f$-divergence are given in \secref{sec:aux:var:properties} of the supplementary material. We show that the bound is tight and reduces to $\fdiv{p(x)}{q(x)}$  when performing a full unconstrained minimization of the bound with respect to $p(z|x)$ (keeping $q(x, z)$ fixed). 

For a latent variable model $p_\theta(x)=\int p_\theta(x|z)p(z)dz$, even if the $f$-divergence between the model and data distribution $\fdiv{p_\theta(x)}{\hat{p}(x)}$ is not computationally tractable, we may form an upper bound 
\begin{align*}
    \fdiv{p_\theta(x)}{\hat{p}(x)} &\leq \fdiv{p_\theta(x|z)p(z)}{\hat{p}(x)q_\phi(z|x)}\\
    &\equiv U(\theta,\phi)
\end{align*}
Provided we choose the quantities in the upper bound appropriately, we can accurately estimate gradients of the bound in order to learn the model parameters $\theta$\footnote{Similar to standard treatments in variational inference (see for example \citet{vae,RezendeMW14}), the variational distribution $q_\phi$ is only being used to tighten the resulting bound and is not a component of the generative model.}. 

The auxiliary variational method \cite{auxvar} and ELBO \cite{vae} are special cases of this objective (see supplementary material). Whilst the bound holds for any $f$-divergence, there can be issues in directly applying this. For example, the corresponding reverse KL bound is
\begin{equation*}
    \KL{p_\theta(x)}{\hat{p}(x)} \leq \KL{p_\theta(x|z)p(z)}{\hat{p}(x)q_\phi(z|x)}
\end{equation*}
For any given $z$, the upper bound has support only on the training data $x_n$, $n=1,\ldots, N$, whereas the model $p_\theta(x|z)$ will typically have larger support. This means that $\int p_\theta(x|z)\log \hat{p}(x)dx$ is ill-defined and the KL divergence (and its gradient) cannot be determined. More generally, calculating the $f$-divergence $\fdiv{p(x)}{q(x)}$ can be problematic when the supports of $p$ and $q$ are different.  To address this and extend the class of latent models and $f$-divergences for which the upper bound can be applied, we make use of the recently introduced Spread Divergence \cite{zhang2020spread}, giving a brief outline below. 

\subsection{Spread $\f$-divergence \label{sec:spread_div}}

For $q(x)$ and $p(x)$ which have disjoint supports, we define new distributions $q(y)$ and $p(y)$ using
\begin{equation*}
p(y) = \int p(y|x)p(x)dx, q(y) = \int p(y|x)q(x)dx
\end{equation*}
where $p(y|x)$ is a "noise" process designed such that $p(y)$ and $q(y)$ have the same support. For example, if we use a Gaussian $p(y|x)=\ndist{y}{x}{\sigma^2}$, then $p(y)$ and $q(y)$ both have support $\mathbb{R}$.

We thus define the \emph{spread $\f$-divergence}
\beq
\sfdiv{p(x)}{q(x)}=\fdiv{p(y)}{q(y)}
\eeq
This satisfies the requirements of a divergence, that is $\sfdiv{p(x)}{q(x)}\geq 0$ and $\sfdiv{p(x)}{q(x)}=0$ if and only if $q(x)=p(x)$.

The auxiliary upper bound can be easily applied to the spread $\f$-divergence
\begin{align} \label{eq:spread_fdiv_bound}
\sfdiv{p(x)}{q(x)}\leq \fdiv{p(y,z)}{q(y,z)}
\end{align}
For example, applying this to the reverse KL objective
\begin{align}
     &\KLs{p_\theta(x)}{\hat{p}(x)} \equiv \KL{p_\theta(y)}{\hat{p}(y)}\nonumber\\
    &\hspace{1.5cm} \leq \KL{p_\theta(y|z)p(z)}{\hat{p}(y)q_\phi(z|y)}
    \label{eq:RKL:spread}
\end{align}
In this case, the spreaded empirical distribution for Gaussian $p(y|x)$ is a mixture of Gaussians
\[
\hat{p}(y)= \int p(y|x)\hat{p}(x)dx = \frac{1}{N\sqrt{2\pi\sigma^2}}\sum_{n=1}^N e^{-\frac{1}{2\sigma^2}\br{y-x_n}^2}
\]
which has support $\mathbb{R}$. Since the spreaded model
\[
p_\theta(y|z) = \int p(y|x)p_\theta(x|z)dx
\]
also has support $\mathbb{R}$, the spread reverse KL divergence $\KLs{p_\theta(x)}{\hat{p}(x)}$ is well defined.

\subsection{Training implicit generative models}

For certain applications (such as image generation) a criticism of latent models $p_\theta(x|z)p(z)$ is that the distribution $p_\theta(x|z)$ may add noise to otherwise "clean" images. For example, if $p_\theta(x|z)=\ndist{x}{\mu_\theta(z)}{\sigma^2}$, then the generated $x$ will add Gaussian noise of variance $\sigma^2$ to the mean image $\mu_\theta(z)$, blurring the image. In practice, many authors fudge this by simply drawing from the model $\delta\br{x-\mu_\theta(z)}p(z)$, without adding on the Gaussian noise. Strictly speaking, this is inconsistent since the model is trained under the assumption of additive Gaussian noise on the observation, but the use of the model differs, assuming no additive noise.  Using the spread divergence, however, we may now directly and consistently define a training objective for 
%
implicit models with deterministic output  
$p_\theta(x,z)=p_\theta(x|z)p(z) = \delta(x - \mu_\theta(z))p(z)$. 
Using spread noise,
\begin{align*}
p_\theta(y|z) &= \int p(y|x)\delta(x - \mu_\theta(z))p(x)dx\\
&= p(y|x=\mu_\theta(z))
\end{align*}
For example, for Gaussian spread noise in one-dimension, we have
\[
p_\theta(y|z) = \ndist{y}{\mu_\theta(z)}{\sigma^2}
\]
which can be used now in the reverse KL bound \eqref{eq:RKL:spread} to train an implicit generative model.
%
%
We thus learn a generative model on the $y$-space, $p_\theta(y)$, by minimizing the $f$-divergence to the noise corrupted data distribution $\hat{p}(y)$. After training we then recover the generative model on $x$-space, $p_\theta(x)$, by taking the mean of our generation network $p_\theta(y|z)$ model as the output.


The upper bound \eqref{eq:RKL:spread} can be estimated through sampling and minimized with respect to $\theta$ and $\phi$ by using the reparameterization trick \cite{vae} and taking gradients. The only difference with standard training (for example stochastic gradient training in the standard VAE) is that an additional outer loop is required in which noisy training points $y$ are sampled from the empirical distribution $\hat{p}(x)$.


Note that for reverse KL training, we require the estimation of the gradient $\triangledown_\theta \int p_\theta (y|z) p(z) \log \hat{p}(y) dy dz $, for which we propose an efficient estimator in  the next section.\\

For $\f$-divergences other than the reverse KL,
\begin{align}
     &\sfdiv{p_\theta(x)}{\hat{p}(x)} \equiv \fdiv{p_\theta(y)}{\hat{p}(y)}\nonumber\\
    &\hspace{1.5cm} \leq \fdiv{p_\theta(y|z)p(z)}{\hat{p}(y)q_\phi(z|y)}
    \label{eq:f:spread}
\end{align}
since the upper bound is expressed as an expectation over $p_\theta(y|z)p(z)$, we can generate $(y, z)$ samples from these distributions and then estimate the bound and take gradients.

\subsection{Gradient approximation}\label{sec:unbiased_gradient}
For the reverse KL upper bound, we need to calculate $\triangledown_\theta \int p_\theta (y|z) p(z) \log \hat{p}(y) dy dz $, where $\hat{p}(y)=N^{-1}\sum_n p(y | x_n)$, i.e. a sum of delta functions (the data distribution) corrupted with a noise process as described in \secref{sec:spread_div}.

Clearly, summing over all points in the dataset to calculate $\hat{p}(y)$ is computationally burdensome. Naively using a minibatch inside the $\log$ to estimate $\hat{p}(y)=M^{-1}\sum_{m\in\mathcal{M}}p(y|x_m)$ results in a biased estimator, which we have found to be detrimental to the optimization procedure in our image generation experiments. Therefore, we propose an different gradient estimator.

We first rewrite the gradient as following (up to addition of a constant):
\begin{align}
\begin{split}
    &\triangledown_\theta \int p_\theta (y|z) p(z) \log \hat{p}(y) dz dy\\
    =&\triangledown_\theta \int p_\theta (y|z) p(z) \sum_n p(n|y) \log p(y|x_n) dz dy \label{eq:gradient_approximation}
\end{split}
\end{align}
Where $p(n|y)=\frac{p(y|x_n)}{\sum_n p(y|x_n)}$ (see \supref{sec:minibatch} for details).
We can now sample a minibatch of index $\mathcal{M}=\{n_1,n_2,...,n_m\}$ from $p(n|y)$ and approximate equation \eqref{eq:gradient_approximation} by
\begin{align}
\begin{split}
    &\triangledown_\theta \int p_\theta (y|z) p(z) \sum_n p(n|y) \log p(y|x_n) dz dy\\
    \approx & \triangledown_\theta \int p_\theta (y|z) p(z) \frac{1}{M}\sum_{m\in\mathcal{M}} \log p(y|x_m) dz dy.
\end{split}
\end{align}

This gradient estimator is unbiased, but we also propose two tricks in \supref{sec:minibatch} to further reduce the computational cost and variance during training. These tricks may bias the estimator, but we find that they work well in practice.

\section{Experiments}

In the following experiments our interest is to demonstrate the applicability of the $\f$-divergence upper bound. The goal of the experiments is not achieving the state-of-the-art image generation results but showing the effectiveness of training with different divergences. Thus the main focus is on training with the reverse KL divergence since this provides a natural "opposite" to training with the forward KL divergence. Throughout, the data is continuous and we use a Gaussian noise process with width $\sigma$ for $p(y|x)$. We take $p(\v{z})$ to be a standard zero mean unit covariance Gaussian (thus with no trainable parameters). Similar to standard VAE training, we use deep networks to parameterize the Gaussian model $p_{\theta}(y|z)=\ndist{y}{\mu_\theta(z)}{\sigma^2}$ and Gaussian variational distribution $q_{\phi} (z|y)=\ndist{z}{\mu_\phi(y)}{\Sigma_\phi(y)}$ for diagonal $\Sigma_\phi(y)$.
Experimentally, we found that running several optimizer steps on $\phi$ whilst keeping $\theta$ fixed is also useful to ensure that the bound is tight when adjusting $\theta$. We therefore use this strategy throughout training. The result that optimizing the auxiliary bound with respect to only $q_\phi(z|y)$ tightens the bound (towards the marginal divergence) is shown in \supref{sec:aux:var:properties}.

\begin{figure}[t]
    \centering{$p_\theta(x)$ samples}

    \centering
    \begin{subfigure}[b]{0.3\textwidth}
        \includegraphics[width=\textwidth]{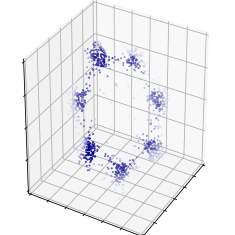}
        \caption{Forward KL}
        \label{fig:toy_fkl}
    \end{subfigure}
    \begin{subfigure}[b]{0.3\textwidth}
        \includegraphics[width=\textwidth]{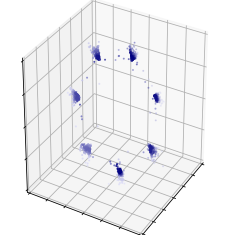}
        \caption{JS divergence}
        \label{fig:toy_js}
    \end{subfigure}
    \begin{subfigure}[b]{0.3\textwidth}
        \includegraphics[width=\textwidth]{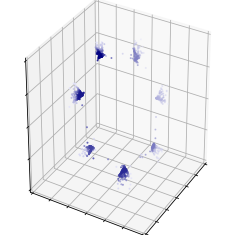}
        \caption{Reverse KL}
        \label{fig:toy_rkl}
    \end{subfigure}

    \centering{Latent space}

    \begin{subfigure}[b]{0.3\textwidth}
        \includegraphics[width=\textwidth]{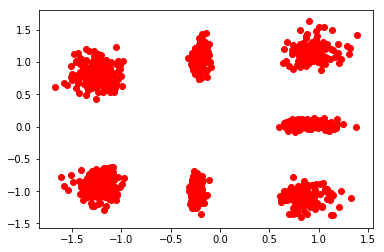}
        \caption{Forward KL}
        \label{fig:toy_fkl_l}
    \end{subfigure}
    \begin{subfigure}[b]{0.3\textwidth}
        \includegraphics[width=\textwidth]{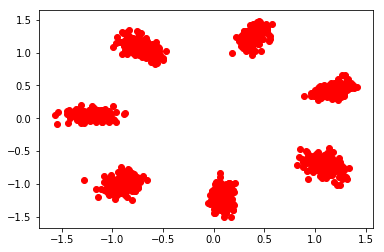}
        \caption{JS divergence}
        \label{fig:toy_js_l}
    \end{subfigure}
    \begin{subfigure}[b]{0.3\textwidth}
        \includegraphics[width=\textwidth]{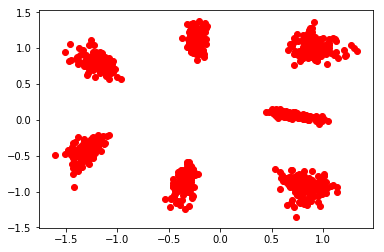}
        \caption{Reverse KL}
        \label{fig:toy_rkl_l}
    \end{subfigure}
    \caption{Toy problem. In (a), (b) and (c) we plot the samples of the model $p_\theta(x)$ trained by three different $\f$-divergences. In (d), (e) and (f) we plot the latent $z$ by sampling from the trained $q_{\phi}(z|x)$ for each datapoint $x$. Note that this is after inverting the noise process, to recover the model on the $x$ space. See also \secref{sec:target}\label{fig:toy}}
\end{figure}
\subsection{Toy problem : Forward KL, Reverse KL and JS training}

The toy dataset, as described by \cite{roth2017stabilizing}, is a mixture of seven two-dimensional Gaussians arranged in a circle and embedded in three dimensional space, see figures: \figref{fig:toy}, \figref{fig:toy_target_dist}. We use 5 hidden layers of 400 units and relu activation function for the mean and variance parameterization in $q_\phi(z|y)$ and mean parameterization in $p_\theta(y|z)$ with a two dimensional latent space $z\in\mathbb{R}^2$.

We use the KL (moment matching) objective $\KL{p(y)}{p_\theta(y)}$, reverse KL (mode seeking) objective $\KL{p_\theta(y)}{p(y)}$ objective, and JS divergence (balance between KL and reverse KL) objective $\JS{p_\theta(y)}{p(y)}$ to train the model. We minimize the corresponding auxiliary $\f$-divergence bounds by gradient descent using the Adam optimizer \citep{adam} with learning rate $10^{-3}$.

To evaluate the bound in each iteration we use a minibatch of size 100 to calculate ${p}^{(B)}(\v{y})$. For each minibatch we draw 100 samples from $p(z)$ and subsequently draw 10 samples from $p_\theta(y|z)$ for each drawn $z$ to generate $(y, z)$ samples. To facilitate training, we anneal the width, $\sigma$ , of the spread divergence throughout the optimization process. This enables the model to feel the presence of other distant modes (high mass regions of $p(y)$), allowing the method to overcome any poor initialization of $(\theta,\phi)$. In this experiment, $\sigma$ is annealed from $1.0$ to $0.1$ using the formula $\sigma=1.0*(0.1^{\frac{\text{current steps}}{\text {total steps}}})$.

We can see in \figref{fig:toy} that the model trained by JS and reverse KL divergence converge to cover each mode in the true generating distribution, and exhibits good separation of the seven modes in the latent space. Even though the reverse KL tends to collapse a model to a single mode, provided the model $p_\theta(y|z)$ is sufficiently powerful, it can correctly capture all the 7 modes.

\begin{figure*}[h]
    \centering
    \begin{subfigure}[b]{0.4\textwidth}
        \includegraphics[width=\textwidth]{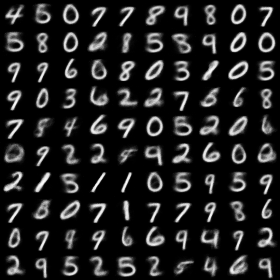}
        \caption{Forward KL}
        \label{fig:fkl_mnist}
    \end{subfigure}
    ~
    \begin{subfigure}[b]{0.4\textwidth}        \includegraphics[width=\textwidth]{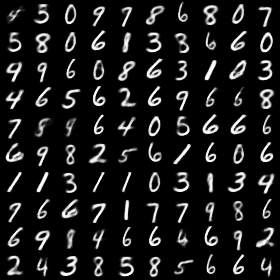}
        \caption{Reverse KL}
        \label{fig:rkl_mnist}
    \end{subfigure}
    \caption{MNIST experiment. (a) Samples from the models trained by forward KL  (b) Samples from the models trained by reverse KL.\label{fig:mnist}}
\end{figure*}

\subsection{MNIST : forward and reverse KL training}
To model the standard MNIST handwritten character dataset \cite{mnist}, we parametrize the mean of $p_\theta(y|z)$ by a neural network and the standard deviation of the spread divergence is fixed to 0.8 for reverse KL training and  0.2 for forward KL training. For $q_\phi(z|y)$ we use a Gaussian with mean and isotropic covariance parameterized by a neural network.
Both networks contain 5 layers, each layer with 512 units and leaky-relu as activation function.  The latent $z$ has dimension 2. We train both forward KL and reverse KL divergence using the $\f$-divergence upper bound. 
For the reverse KL experiment, we first initialize the model by training using the forward KL objective for 20 epochs to prevent getting a sub-optimal solution (e.g. mode collapsing). After that, we train the model using both forward KL and reverse KL objective for additional 40 epochs. For the reverse KL training, we interleave each $\theta$ update (learning rate $5e^{-5}$ with SGD) with 20 $\phi$ updates (learning rate $10^{-4}$ with Adam), the batch size is 100 in both cases. We also use the gradient approximation method discussed in \secref{sec:unbiased_gradient}, also see \supref{sec:minibatch}. 

In \figref{fig:mnist}, we show the samples from the two trained models using the same latent codes. As we can see, the samples generated by reverse KL model are sharper.


\subsection{CelebA: forward and reverse KL training}

We pre-process CelebA \citep{liu2015faceattributes} dataset (with 50000 images) by first taking 140x140 center crops and then resizing to 64x64. Pixel values were then rescaled to lie in $[0,1]$. The architectures of the convolutional encoder $q_\phi(z|y)$ and deconvolutional decoder $p_\theta(y|z)$ (with fixed noise) are given in the supplementary material \secref{sec:arch}.
The standard deviation of the spread divergence is 0.2 for the KL and 1.0 for the RKL. We first train the model using KL divergence for 60 epochs as initialization and then train for 6 additional epoch for both pure forward KL and reverse KL. In order to ensure that the bound remained tight, we interleave each $\theta$ update (learning rate $10^{-7}$ with RMSprop) with 20 $\phi$ updates (learning rate $10^{-4}$ with Adam), the batch size is 100 in both cases.

\begin{figure*}[h]
    \centering
    \begin{subfigure}[b]{0.4\textwidth}
        \includegraphics[width=\textwidth]{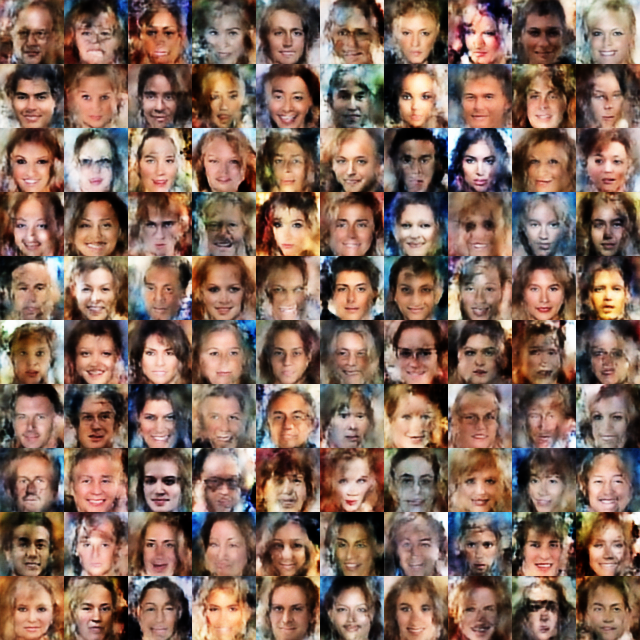}
        \caption{Forward KL}
        \label{fig:gull}
    \end{subfigure}
    ~
    \begin{subfigure}[b]{0.4\textwidth}
        \includegraphics[width=\textwidth]{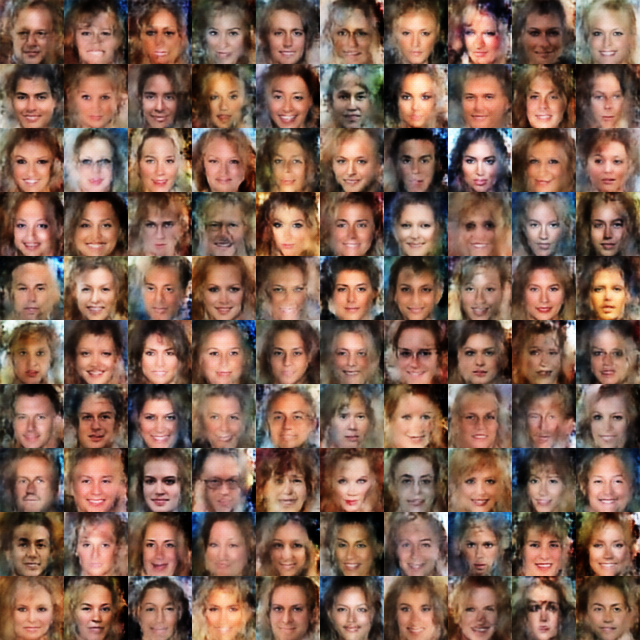}
        \caption{Reverse KL}
        \label{fig:tiger}
    \end{subfigure}
    \caption{CelebA experiment. Image samples from the trained models $p_\theta(x)$. After VAE initialization, we continued training for an additional epoch with (a) pure forward KL and (b) the reverse KL divergence.
    \label{fig:celebA}}
\end{figure*}

In \figref{fig:celebA} we show samples from the two trained models using the same latent codes. As we can see, the impact of the reverse KL term in training is significant, resulting in less variability in pose, but sharper images.  This is consistent with the "mode-seeking" behavior of the reverse KL objective.

\subsection{Comparison to Fenchel conjugate lower bound \label{sec:fgan_exp}}

As discussed in \secref{sec:estmation_f}, a different approach to minimizing the $\f$-divergence is used in \cite{nowozin2016f}, utilizing a variational \emph{lower bound} to the $\f$-divergence:
\begin{align}
\begin{split}
&\fdiv{p(x)}{p_\theta(x)}\\& \geq \sup_{T \in \mathcal{T}} \big( \mathbb{E}_{x \sim p(x)}[T(x)] - \mathbb{E}_{x \sim p_\theta(x)}[f^*(T(x))] \big)
\end{split}
\label{eq:fgan}
\end{align}
Here $f^*$ is the Fenchel conjugate and $\mathcal{T}$ is any class of functions that respects the domain of $f^*$. After parameterizing $T=g_f(V_\phi)$ (where $g_f:\mathbb{R}\rightarrow dom_{f^*}$ and $V_\phi$ is an unconstrained parametric function) and $p_\theta(x)$, the optimization scheme is then to alternately tighten (i.e. increase) the bound through changes to $\phi$ and then lower the bound through changes to $\theta$, see \figref{fig:bounds}. This is of interest because the GAN objective \citep{gan_paper} can be seen as a specific instance of this scheme. We acknowledge that the $f$-GAN principally grounds GANs in a wider class of techniques, and is not necessarily intended as a scheme for minimizing an $\f$-divergence. However, it is natural to ask whether our auxiliary upper bound or the Fenchel-conjugate lower bound gives different results when used to minimize the $\f$-divergence for a similar complexity of parameter space $(\theta,\phi)$.

To compare the two methods we fit a univariate Gaussian $p_\theta(x)$ to data generated from a mixture of two Gaussians through the minimization of various $\f$-divergences. See the supplementary material for details. For the $\f$-GAN lower bound we use a network with two hidden layers of size 64 for $V_{\phi}(x)$. For our upper bound we use a network with two hidden layers of size 50 to parameterize $q_\phi(z|x)$ and set $p_\theta(x, z)$ to be a bivariate Gaussian, so that it marginalizes to a univariate Gaussian as required. The upper and lower bound methods have a similar number of free parameters ($q_\phi$ has fewer hidden units but more outputs than $V_\phi$). The two methods result in broadly similar Gaussian fits, see \tabref{table:fgan_comparison}. In general, minimizing the upper bound results in a slightly superior fit compared to the $\f$-GAN method \citep{nowozin2016f} in terms of proximity to the true minimal $\f$-divergence fit and proximity of the bound value to the true divergence. Additionally we find that minimizing our upper bound is computationally more stable than the optimization procedure required for $\f$-GAN training (simultaneous tightening and lowering of the bound -- see supplementary material \ref{sec:fgan:comp}).



\begin{table}[t]

\centering
  \caption{Learned Gaussian parameters to fit a mixture of two Gaussians using forward KL, reverse KL and Jensen-Shannon divergence. $p_{*}(x)$ is the optimal Gaussian fitted to minimize the exact divergence. $p_{UB}(x)$ is the optimal Gaussian fitted to minimize our auxiliary upper bound on the divergence. $p_{LB}(x)$ is the optimal Gaussian fitted to minimize the Fenchel-conjugate lower bound on the divergence.\label{table:fgan_comparison}}
  \begin{tabular}{llll}

    \toprule
    & KL & rev-KL & J-S\\
    \midrule
    \midrule
    $\fdiv{p(x)}{p_*(x)}$ & 0.21 & 0.18 & 0.05 \\
    \midrule
    $\fdiv{p(x)}{p_{LB}(x)}$ & 0.32 & 0.25 & 0.23 \\
    $\fdiv{p(x)}{p_{UB}(x)}$ & \textbf{0.21} & \textbf{0.23} & \textbf{0.15} \\
    \midrule
    \midrule
    $\mu^*$ & 1.70 & 1.85 & 1.76 \\
     \midrule
     $\hat{\mu}_{LB}$ & 1.71 & 1.73 & 1.70 \\
     $\hat{\mu}_{UB}$ & 1.71 & \textbf{1.76} & 1.70 \\
     \midrule
     \midrule
    $\sigma^*$ & 0.62 & 0.57 & 0.60 \\
     \midrule
     $\hat{\sigma}_{LB}$ & 0.46 & 0.45 & 0.24 \\
     $\hat{\sigma}_{UB}$ & \textbf{0.62} & \textbf{0.65} & \textbf{0.33} \\
    \bottomrule
  \end{tabular}
\vskip -0.1in

\end{table}
\vskip -1in

\section{Related work}

The \textit{Auxiliary Variational Method} \citep{auxvar} uses an auxiliary space to minimize the joint KL divergence in order to minimize the marginal KL divergence. We extend this method to the more general class of $\f$-divergences.

Compared to \textit{Variational Auto-Encoders} \citep{vae}, 
our method is a way to train an identical class of generative and variational models, but with a class of different optimization objectives based on $\f$-divergences. Since the VAE optimization scheme is a variational method of maximizing the likelihood, it is similar to our scheme with the choice of minimizing the forward KL divergence, which is also a variational form of maximum likelihood. Both methodologies use sampling to estimate a variational bound which can be differentiated through the use of the reparameterization trick.

In \textit{R\'enyi divergence variational inference }\citep{li2016renyi}, a variational approximation of log-likelihood is proposed based on the  R\'enyi divergence. However, our joint upper bound is an estimator of $\f$-divergence in marginal data space, it only relates to maximum likelihood learning when we use KL divergence.

In \textit{Auxiliary Deep Generative Models} \citep{maaloe}, a VAE is extended with an auxiliary space. This allows a richer variational distribution to be learned, with the correlation between latent variables being pushed to the auxiliary space to keep the calculation tractable. This, similarly to our method, utilizes the general auxiliary variational method \cite{auxvar}, but is focused on making VAEs more powerful rather than providing different optimization schemes.

In the $\f$-GAN \citep{nowozin2016f} methodology, an interesting connection is made between the GAN training objective and a lower bound on the $\f$-divergence.
The authors conclude that using different divergences leads to largely similar results, and that the divergence only has a large impact when the model is "under-powered". However, that conclusion is somewhat at odds with our own, in which we find that the (upper bound on) different divergences gives very different model fits. Indeed, others have reached a similar conclusion: the reverse KL divergence is optimized as a GAN objective in \cite{sonderby2016amortised}, demonstrating that it is effective in the task of image super-resolution. A variety of different generator objectives for GANs are used in \cite{poole}, with some divergence objectives exhibiting the "mode-seeking" behavior we have observed.

In \cite{mohamed2016learning}, the authors demonstrate an alternative approach to train $\mathnormal{f}$-divergence $\fdiv{p(x)}{q(x)} = \int q(\v{x})f\br{\frac{p(\v{x})}{q(\v{x})}} d \v{x}$ by directly estimating the density ratio $\frac{p(x)}{q(x)}$. This method makes a connection to GANs: the discriminator is trained to approximate the ratio and the generator loss is designed based upon different choices of $\mathnormal{f}$-divergence (see the supplementary material \secref{sec:cpe} for details). We thereby recognize there are three different tractable estimations of the $\mathnormal{f}$-divergence: 1. ratio estimation in the marginal space 2. Fenchel conjugate lower bound ($\mathnormal{f}$-GAN) and 3. the variational joint upper bound (introduced by our paper).

Ratio estimation by classification has also been extended to minimize the KL-divergence in the joint space \citep{huszar2017variational}. Similarly, \textit{Bi-directional GAN}  \citep{donahue2016adversarial} and \textit{Ali-GAN} \citep{dumoulin2016adversarially} augment the GAN generator with an additional inference network.
Although these models focus on similar training objectives to our own, the purpose of using the joint space is different to that of our approach. Our method uses the joint distribution to create an upper bound in order to estimate the $\mathnormal{f}$-divergence in the marginal space; the latent representation is automatically achieved. In contrast, all three methods mentioned above expand the original space to a joint space just for learning the latent representation, the divergence is estimated by either ratio estimation or GAN approaches. Additionally, they only minimize the target divergence only at the limit of an optimal discriminator (or in the nonparametric limit, see \cite{gan_paper} and \cite{mescheder2017adversarial}), which may cause instability in the GAN training process \citep{arjovsky2017towards}.

\section{Conclusion}

We introduced an upper bound on $\f$-divergences, based on an extension of the auxiliary variational method. The approach allows variational training of latent generative models in a much broader set of divergences than previously considered. We showed that the method requires only a modest change to the standard VAE training algorithm but can result in a qualitatively very different fitted model.  For our low dimensional toy problems, both the forward KL and reverse KL can be effective in learning the model. However, for higher dimensional image generation, compared to standard forward KL training (VAE), training with the reverse KL tends to focus much more on ensuring that data is generated with high fidelity around a smaller number of modes.  The central contribution of our work is to facilitate the application of more general $\f$-divergences to training of probabilistic generative models with different divergences potentially giving rise to very different learned models. 
\nocite{langley00}

\bibliography{example_paper}
\bibliographystyle{icml2019}

\appendix

\onecolumn
\appendix

\newpage

\section*{Supplementary Material}

\section{Properties of the Auxiliary Variational Method\label{sec:aux:var:properties}}

Here we give a property of the auxiliary bound for $\f$-divergences with differentiable $\f$; this covers most $\f$ of interest, and the argument extends to those $\f$ which are piecewise differentiable. Then for the particular case of the reverse Kl divergence we give a simpler proof of this property as well as two additional properties (which do not hold for general $\f$).

\subsection{$\fdiv{p(x)}{q(x)}$} \label{sec:fdiv_marg}

For differentiable $\f$ we claim that when we fully optimize the auxiliary $\f$-divergence w.r.t $p(\v{z}|\v{x})$, this is the same as minimizing the $\f$-divergence in the $\v{x}$ space alone.

Let's first fix $q(\v{x}, \v{z})$ and find the optimal $p(\v{z}|\v{x})$ by taking the functional derivative of the auxiliary $\f$-divergence

\begin{align}
    \frac{\delta}{\delta p(z|x)} \fdiv{P}{Q} &= \frac{\delta}{\delta p(z|x)} \int q(x', z')f\br{\frac{p(z'|x')p(x')}{q(x', z')}} dx' dz' \\
    &= q(x, z) f'\br{\frac{p(z|x)p(x)}{q(x, z)}} \frac{p(x)}{q(x, z)} \\
    &= p(x)f'\br{\frac{p(z|x)p(x)}{q(z|x)q(x)}}
\end{align}

At the minimum this will be equal to 0 (plus a constant Lagrange multiplier that comes from the constraint that $p(z|x)$ is normalized). Since $f'$ is not constant (if it is then the $\f$-divergence is a constant), this then implies that the argument of $f'$ must be constant in $z$. This implies that optimally $p(z|x)=q(z|x)$. Plugging this back into the $\f$-divergence, it reduces to simply $\fdiv{p(x)}{q(x)}$

Hence, we have shown
\beq
\min_{p(\v{z}|\v{x})}\fdiv{p(x, z)}{q(x, z)}= \fdiv{p(x)}{q(x)}
\eeq

Since the assumption is that $\fdiv{p(x)}{q(x)}$ is not computationally tractable, this means that, in practice, we need to use a suboptimal $p(\v{z}|\v{x})$, restricting $p(\v{z}|\v{x})$ to a family $p_\theta(\v{z}|\v{x})$ such that the joint $\f$-divergence is computationally tractable.

\subsection{Relation to $\KL{q(x)}{p(x)}$}

For the particular case of the reverse KL divergence we also provide this more straightforward proof.

Again, the claim is that when we fully optimize the auxiliary KL divergence w.r.t $p(\v{z}|\v{x})$, this is the same as minimizing the KL in the $\v{x}$ space alone.

Let's first fix $q(\v{x}, \v{z})$ and find the optimal $p(\v{z}|\v{x})$. The divergence is
\begin{align} \label{eq:4}
    \begin{split}
        \KL{q(\v{x},\v{z})}{p(\v{x},\v{z})} &=- \int q(\v{z}|\v{x}) q(\v{x}) \log  p(\v{z}|\v{x})dxdz + \const \\
        &= \int q(\v{x}) \KL{q(\v{z}|\v{x})}{p (\v{z}|\v{x})}dx +\const
    \end{split}
\end{align}

Since we are taking a positive combination of KL divergences, this means that, optimally, $p(\v{z}|\v{x})=q(\v{z}|\v{x})$. Plugging this back into the KL divergence, the KL reduces to simply
\beq
\KL{q(\v{x})}{p(\v{x})}
\eeq
Hence, we have shown
\beq
\min_{p(\v{z}|\v{x})}\KL{q(\v{x},\v{z})}{p(\v{z}|\v{x})p(\v{x})}= \KL{q(\v{x})}{p(\v{x})}
\eeq

\subsection{Independence $p(\v{z}|\v{x})=p(\v{z})$}

Also for the particular case of the reverse KL divergence we can derive a result from the assumption that the auxiliary variables are independent of the observations and the prior $p(\v{z}|\v{x}) = p(\v{z})$. We have
\begin{equation}
    \begin{split}
        \KL{q(\v{x}, \v{z})}{p(\v{x},\v{z})} & = \int  q(\v{x},\v{z}) \log  q(\v{x},\v{z})dxdz - \int q(\v{x},\v{z}) \log [p(\v{z}) p(\v{x})]dxdz \\
                              & = \KL{q(\v{z})}{p(\v{z})}  + \int  q(\v{z})\KL{q(\v{x}|\v{z})}{p(\v{x})}dz
    \end{split}
\end{equation}
Optimally, therefore, we set $p(\v{z})=q(\v{z})$, which gives the resulting expression
\beq
\int_\v{z} q(\v{z})\KL{q(\v{x}|\v{z})}{p(\v{x})}
\eeq
Since we are still free to set $q(\v{z})$, we should optimally set $q(\v{z})$ to place all its mass on the $\v{z}$ that minimizes
\beq
\KL{q(\v{x}|\v{z})}{p(\v{x})}
\eeq
In other words, the assumption of independence $p(\v{z}|\v{x})=p(\v{z})$ implies that method is no better than computing each $\KL{q(\v{x}|\v{z})}{p(\v{x})}$ and then choosing the single best model $q(\v{x}|\v{z})$.

%
%
%
%
%
%
%

\subsection{Factorizing $q(\v{x},\v{z})=q(\v{x})q(\v{z})$}

Again for the reverse KL divergence, under the independence assumption $q(\v{x},\v{z})=q(\v{x})q(\v{z})$, it is straightforward to show that
\beq
\KL{q(\v{x}, \v{z})}{p(\v{x},\v{z})} = \KL{q(\v{x})}{p(\v{x})}
\eeq
In the case that $q(\v{x})$ for example is a simple Gaussian distribution, this means that the independence assumption does not help enrich the complexity of the approximating distribution.

\subsection{Relation to the ELBO}


The reverse KL divergence in joint space: $\KL{q(\v{x}, \v{z})}{p(\v{x},\v{z})}$ is equivalent to using the ELBO to lower bound $\log p(x)$ in $\KL{q(x)}{p(x)}$:

\begin{equation}
    \begin{split}
\KL{q(\v{x})}{p(\v{x})}&=\int q(x)(\log q(x)-\log p(x))dx\\
&\leq\int q(x)\br{\log q(x)-\underbrace{\int q(z|x)(\log p(x,z)-\log q(z|x))}_{\text{ELBO}}dz}dx\\
&=\int q(x)q(z|x)(\log q(x)q(z|x)-\log p(x,z))dxdz\\
&=\KL{q(\v{x}, \v{z})}{p(\v{x},\v{z})}
    \end{split}
\end{equation}

\section{Reverse KL gradient approximation}\label{sec:minibatch}
For the reverse KL upper bound, we need to calculate $\triangledown_\theta \int p_\theta (y|z) p(z) \log \hat{p}(y) dy dz $, where $p(y)=N^{-1}\sum_n p(y | x_n)$, let us assume that $p_\theta(y|z)$ and $p(y|x_n)$ are spherical Gaussians with the same variance $\sigma^2$ (which is the case in our experiments)

\begin{align}
    \triangledown_\theta &\int p_\theta (y|z) p(z) \log \hat{p}(y) dy dz \\ & = \triangledown_\theta \int \frac{1}{\br{2 \pi \sigma^2}^{D/2}} \exp \br{-\frac{1}{2 \sigma^2}(y - \mu_\theta(z))^2} p(z) \log \bigg(N^{-1} \sum_n p(y|x_n)\bigg) dy dz\\
    &= \int p(\epsilon) p(z) \triangledown_\theta \log \br{ \frac{1}{N\br{2 \pi \sigma^2}^{D/2}} \sum_n \exp \br{-\frac{1}{2 \sigma^2} \big(\mu_\theta(z) + \sigma \epsilon - x_n\big)^2}} d\epsilon dz\\
    &= \int p(\epsilon) p(z) \frac{\sum_n \triangledown_\theta\big(-\frac{1}{2\sigma^2}(\mu_\theta(z) + \sigma \epsilon - x_n)^2\big) \exp \br{-\frac{1}{2 \sigma^2} \big(\mu_\theta(z) + \sigma \epsilon - x_n\big)^2}}{\sum_n \exp \br{-\frac{1}{2 \sigma^2} \big(\mu_\theta(z) + \sigma \epsilon - x_n\big)^2}}  d\epsilon dz+\const \\
    &= \int p(\epsilon) p(z) p(n|\epsilon, z)\sum_n \triangledown_\theta \big(-\frac{1}{2\sigma^2}(\mu_\theta(z) + \sigma \epsilon - x_n)^2\big)   d\epsilon dz+\const\\
    &=\triangledown_\theta \int p_\theta (y|z) p(z)  \sum_n p(n|y) \log p(y|x_n) dy dz +\const
    \label{eq:log_gradient}
\end{align}
Where we have used the reparametrization $y = \mu_\theta(z) + \sigma \epsilon$ and $p(\epsilon)=N(0, I)$, and then noticed that we can define 

\begin{equation}
    p(n|y)=p(n|\epsilon, z) := \frac{\exp \br{-\frac{1}{2 \sigma^2} \big(\mu_\theta(z) + \sigma \epsilon - x_n\big)^2}}{\sum_n \exp \br{-\frac{1}{2 \sigma^2} \big(\mu_\theta(z) + \sigma \epsilon - x_n\big)^2}}=\frac{p(y|x_n)}{\sum_n p(y|x_n)}
\end{equation}
which is a softmax over the square distance (with a scaling) between $y$ and $x_n$.

We can now get an unbiased estimator for this gradient \eqref{eq:log_gradient} if we can generate samples from $p(n|\epsilon, z)$.

\textbf{Computation reduction}
The computationally expensive part of calculating $p(n|\epsilon, z)$ is the normalizer, which requires summing over all data points. Given that typically the $x$-space will be high dimensional in practice we consider a dimensionality reduction technique to speed up computing the square distance between $y$ and $x_n$.

We use Principal Components Analysis (PCA) to project the $x$-space to a much lower dimensional space. PCA is an appropriate choice as it maximizes the variance preserved by the lower dimensional projections, whilst minimizing the square distance between the reconstructions and the original data. Note that the PCA projection matrix, $U$, is learned once on the input data $\{x_n\}$.

So we approximate

\begin{equation}
    p(n|\epsilon, z) \approx q(n|\epsilon, z) := \frac{\exp \br{-\frac{1}{2 \sigma^2} \big(U^T(\mu_\theta(z) + \sigma \epsilon) - U^T(x_n)\big)^2}}{\sum_n \exp \br{-\frac{1}{2 \sigma^2} \big(U^T(\mu_\theta(z) + \sigma \epsilon) - U^T(x_n)\big)^2}}\label{eq:softmax}
\end{equation}

We can now get an (approximate) unbiased estimator for \eqref{eq:log_gradient} by sampling $\epsilon, z$ and then $n \sim q(n|\epsilon, z)$.

\begin{equation}\label{eq:softmax_approx}
    \triangledown_\theta \int p_\theta (y|z) p(z) \log \hat{p}(y) dy dz \approx  -\frac{1}{S\sigma^2} \sum_{s=1}^S \bigg[ \triangledown_\theta \mu_\theta(z^{(s)})\frac{1}{T}\sum_{t=1}^T \big( \mu_\theta(z^{(s)}) + \sigma \epsilon^{(s)} - x_{(n_s^{(t)})} \big) \bigg]
\end{equation}

Where $z^{(s)} \sim p(z)$, $\epsilon^{(s)} \sim p(\epsilon)$, and for each s we sample $n_s^{(t)} \sim q(n|\epsilon^{(s)}, z^{(s)})$. In the experiments on MNIST/CelebA, we sample $T=30$ to form the approximation, and the PCA dimension is 50/100.

Using this approximation, the computation cost of the normalizer in \eqref{eq:softmax_approx}  scales with the number of the data points, but we have found this not to be an issue in practice when using the PCA projection. For a very large dataset this could be problematic though. In this case other minibatch methods could be used to approximate this normalizer, such as \cite{ruiz2018augment} and \cite{botev2017complementary}, which we leave to further work.

\textbf{Variance reduction}
To reduce the variance in the softmax sampling, we approximate \eqref{eq:softmax} by:
\begin{equation}
    q(n|\epsilon, z) \approx  \frac{\exp \br{-\frac{1}{\mathcal{T}} \big(U^T(\mu_\theta(z)) - U^T(x_n)\big)^2}}{\sum_n \exp \br{--\frac{1}{\mathcal{T}} \big(U^T(\mu_\theta(z)) - U^T(x_n)\big)^2}}\label{eq:softmax}
\end{equation}
Where $\mathcal{T}$ is the temperature. When $\frac{1}{\mathcal{T}}\rightarrow 0$, the sample from softmax distribution is the index of the nearest neighbourhood  in square distance. In all experiments, we set $\mathcal{T}=10$.

Note that by using these two tricks, the estimator is no longer unbiased anymore, but we found they work well in practice.



\begin{wrapfigure}{R}{0.28\textwidth}
\includegraphics[width=1\textwidth]{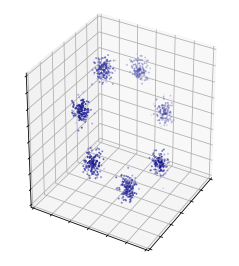}
\caption{Target distribution of the toy problem, from \cite{roth2017stabilizing}}
\label{fig:toy_target_dist}
\end{wrapfigure}

\section{Target distribution of the toy problem\label{sec:target}}

We train on the toy dataset described by \cite{roth2017stabilizing}, which is a mixture of seven two-dimensional Gaussians arranged in a circle and embedded in three dimensional space, see \figref{fig:toy_target_dist}.
The standard deviation of the each Gaussian is 0.05.

\section{Network Architecture\label{sec:arch}}

Both encoder and decoder used fully convolutional architectures with 5x5 convolutional filters and used vertical and horizontal strides 2 except the last deconvolution layer we used stride 1. Here $\text{Conv}_k$ stands for a convolution with $k$ filters, $\text{DeConv}_k$ for a deconvolution
with k filters, BN for the batch normalization \cite{ioffe2015batch}, ReLU for the rectified linear units, and $\text{FC}_k$ for the fully connected
layer mapping to $R^k$.
\begin{align*}
x\in R^{64\times64\times3}&\rightarrow \text{Conv}_{128}\rightarrow\text{BN}\rightarrow\text{Relu}\\
&\rightarrow \text{Conv}_{256}\rightarrow\text{BN}\rightarrow\text{Relu}\\
&\rightarrow \text{Conv}_{512}\rightarrow\text{BN}\rightarrow\text{Relu}\\
&\rightarrow \text{Conv}_{1024}\rightarrow\text{BN}\rightarrow\text{Relu}\rightarrow\text{FC}_{64}
\end{align*}

\begin{align*}
z\in R^{64}&\rightarrow \text{FC}_{8\times 8\times 1024}\\ &\rightarrow\text{DeConv}_{512}\rightarrow\text{BN}\rightarrow\text{Relu}\\
&\rightarrow \text{DeConv}_{256}\rightarrow\text{BN}\rightarrow\text{Relu}\\
&\rightarrow \text{DeConv}_{128}\rightarrow\text{BN}\rightarrow\text{Relu}\\
&\rightarrow \text{DeConv}_{64}\rightarrow\text{BN}\rightarrow\text{Relu}\rightarrow\text{DeConv}_{3}
\end{align*}

\section{$\f$-GAN comparison\label{sec:fgan:comp}}

\begin{figure}[h]
\centering
\includegraphics[width=0.5\textwidth]{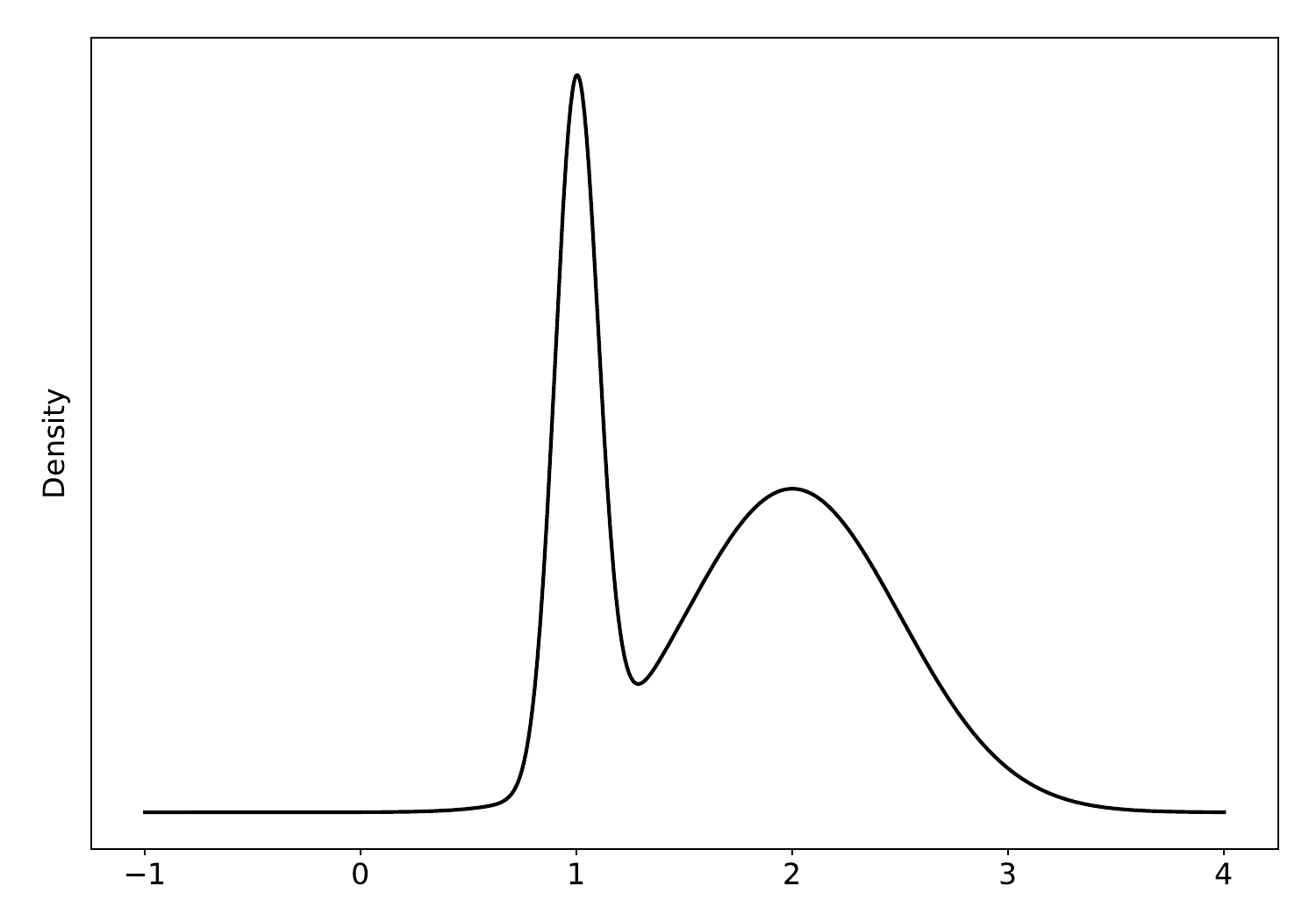}
\caption{Mixture of two Gaussians, $0.3\mathcal{N}_1 + 0.7\mathcal{N}_2$ where $\mathcal{N}_1 = \mathcal{N}(\mu_1=1, \sigma_1=0.1)$ and $\mathcal{N}_2 = \mathcal{N}(\mu_2=2, \sigma_2=0.5)$}
\label{fig:mog_fgan}
\end{figure}

The mixture of Gaussians we attempt to fit a univariate Gaussian to is plotted in Figure \ref{fig:mog_fgan}.

\begin{figure}[t]
    \centering
    \begin{subfigure}[b]{0.49\textwidth}
        \includegraphics[width=\textwidth]{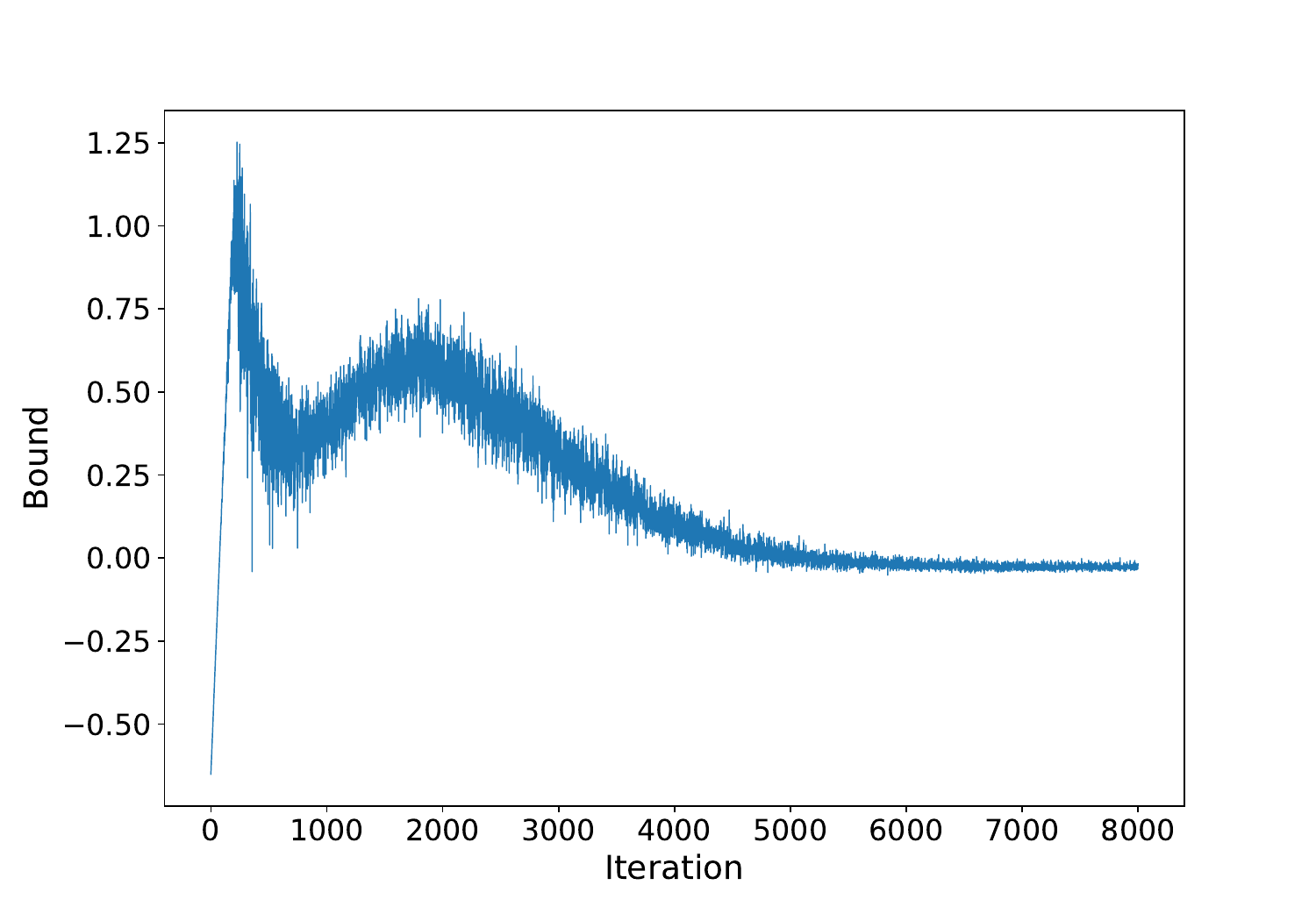}
        \caption{Forward KL lower bound}
    \end{subfigure}
    \hfill
    \begin{subfigure}[b]{0.49\textwidth}
        \includegraphics[width=\textwidth]{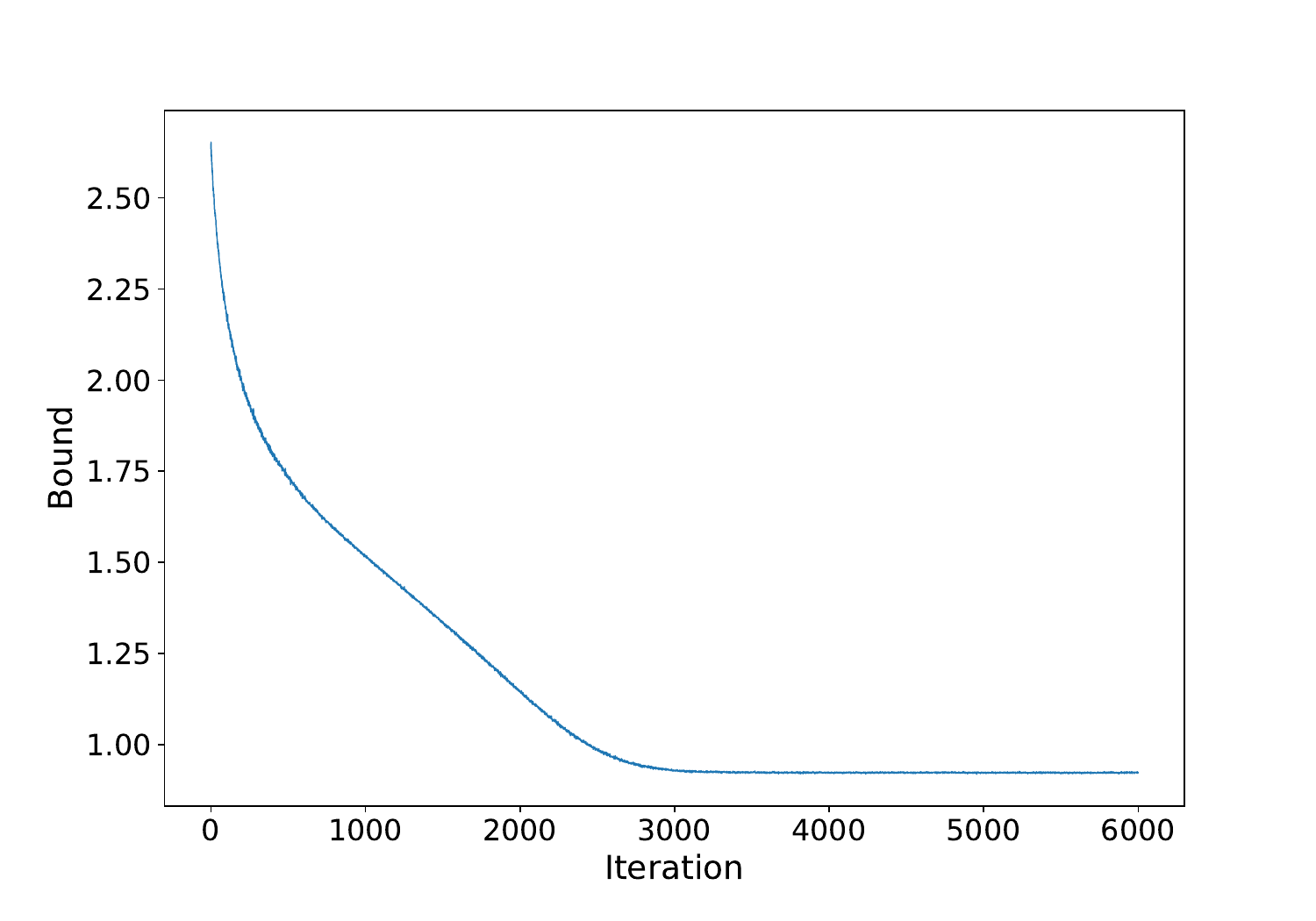}
        \caption{Forward KL upper bound}
    \end{subfigure}

    \begin{subfigure}[b]{0.49\textwidth}
        \includegraphics[width=\textwidth]{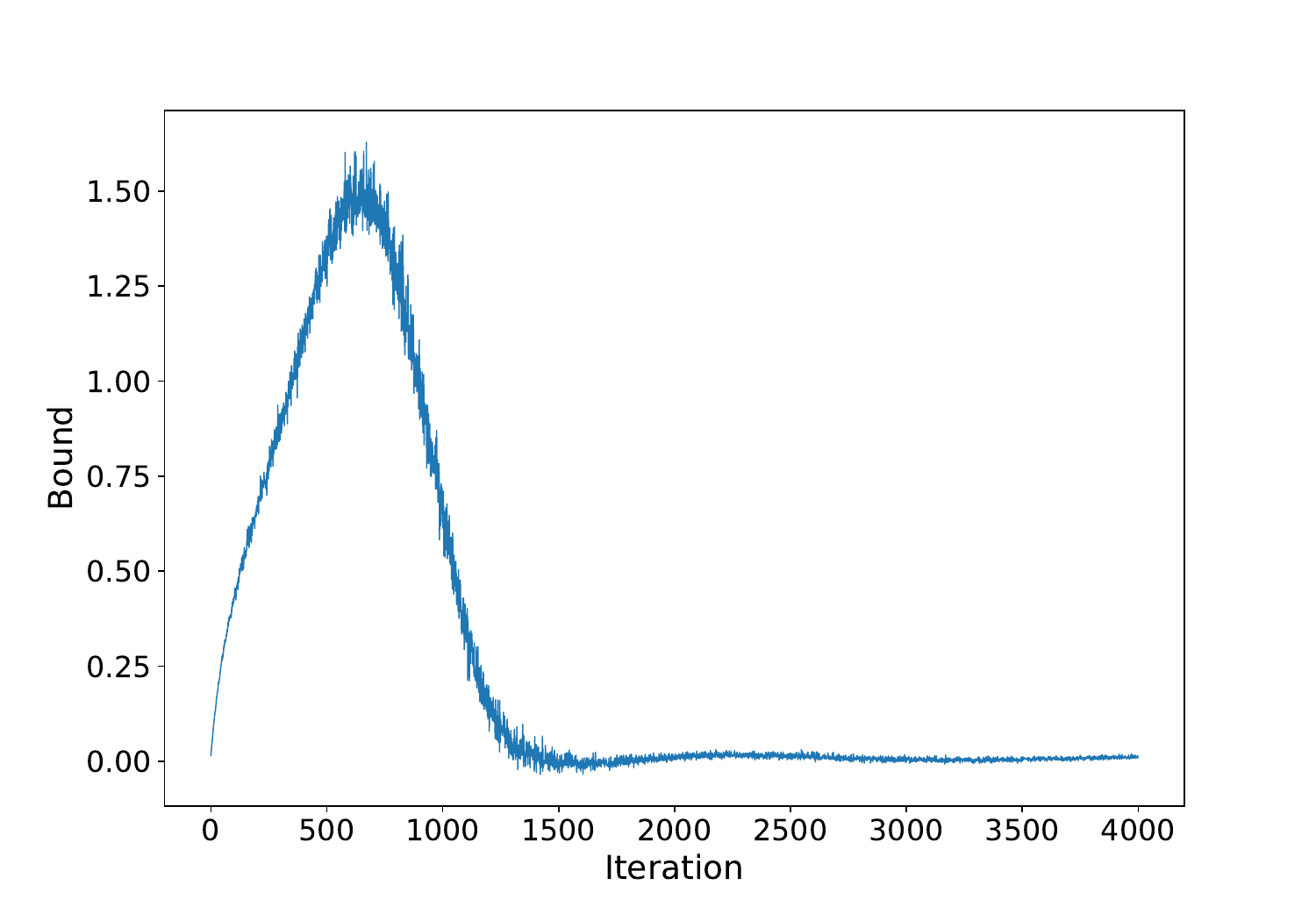}
        \caption{Reverse KL lower bound}
    \end{subfigure}
    \hfill
    \begin{subfigure}[b]{0.49\textwidth}
        \includegraphics[width=\textwidth]{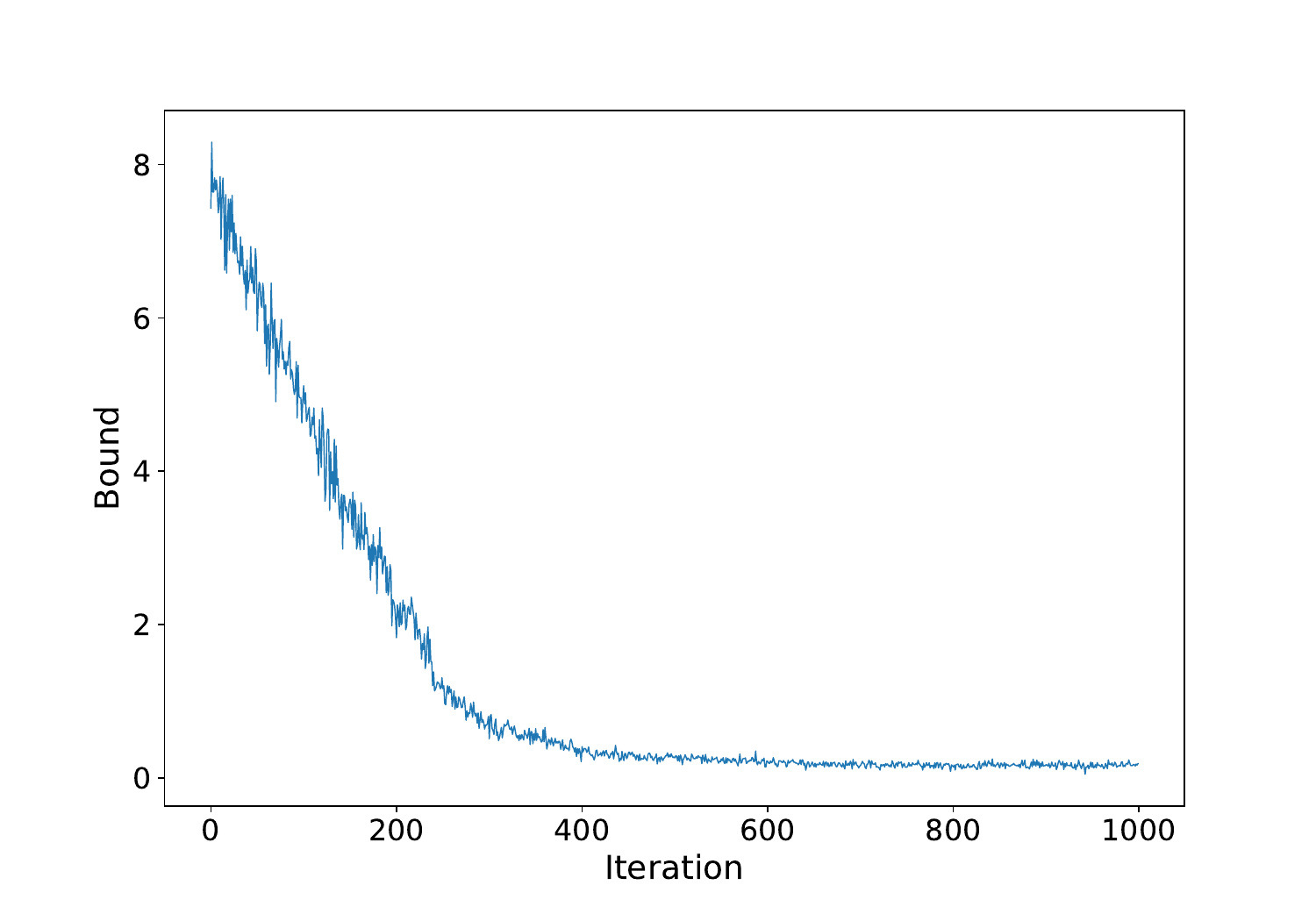}
        \caption{Reverse KL upper bound}
    \end{subfigure}

    \begin{subfigure}[b]{0.49\textwidth}
        \includegraphics[width=\textwidth]{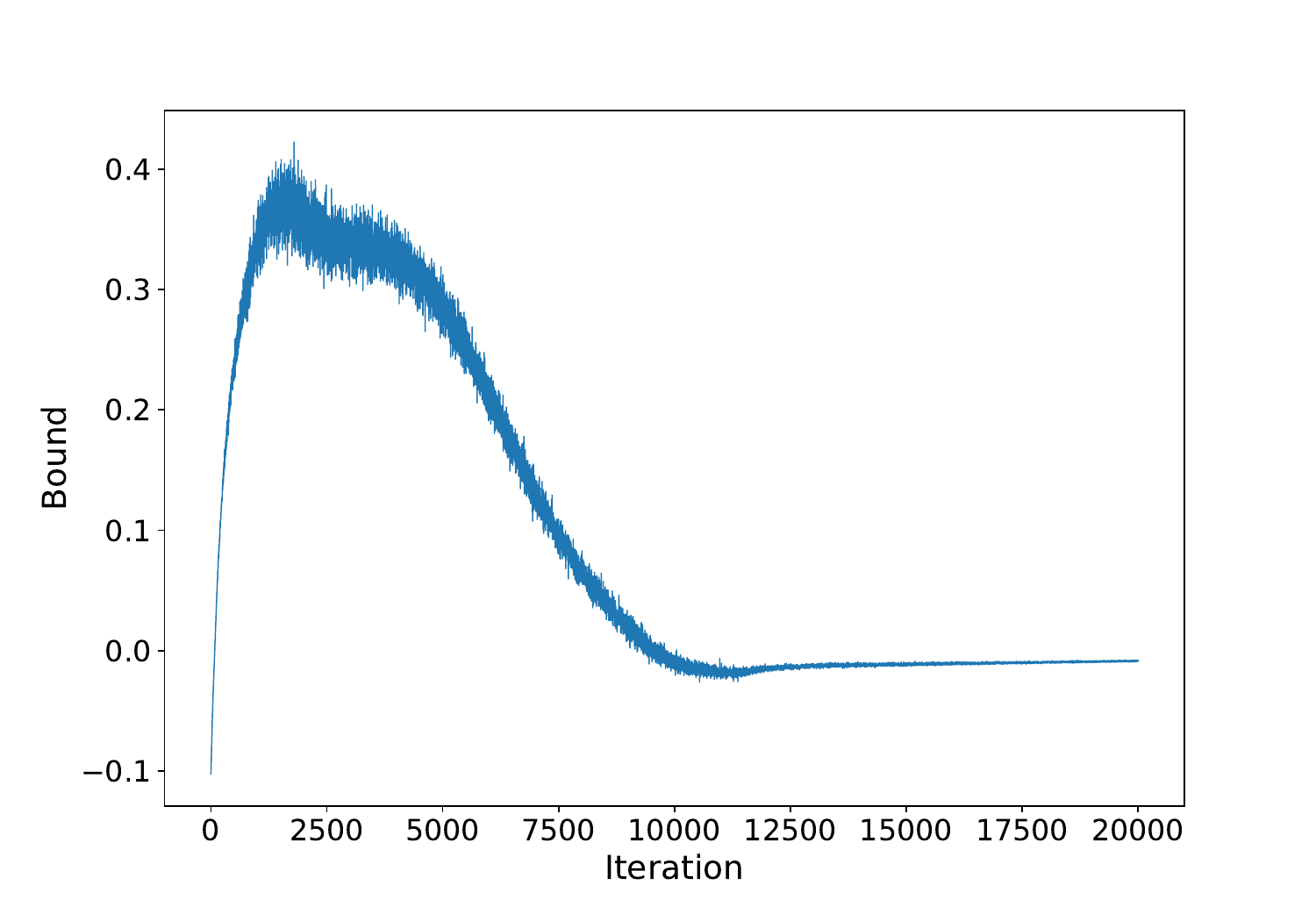}
        \caption{Jensen-Shannon lower bound}
    \end{subfigure}
    \hfill
    \begin{subfigure}[b]{0.49\textwidth}
        \includegraphics[width=\textwidth]{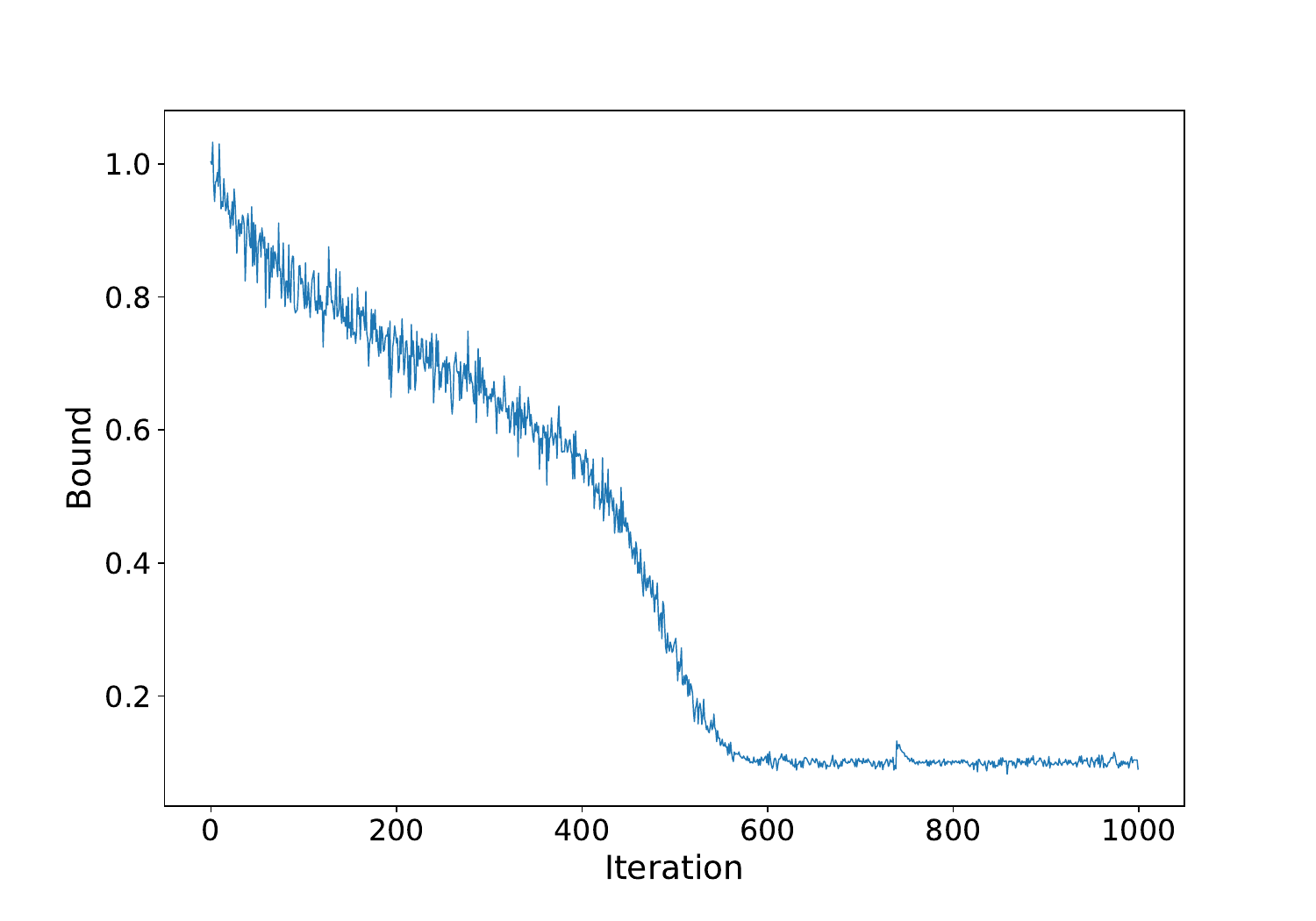}
        \caption{Jensen-Shannon upper bound}
    \end{subfigure}

    \caption{Training runs for fitting a univariate Gaussian to a mixture of two Gaussians by minimizing a variety of $\f$-divergences. On the left we train using the lower bound, on the right with our upper bound. \label{fig:fgan_training}}
\end{figure}

We plot the lower and upper bounds during training in Figure \ref{fig:fgan_training}. We can see the upper bound is generally faster to converge and less noisy. It also a consistently decreasing objective, whereas the variational lower bound fluctuates higher and lower in value throughout the training process.

\section{Class Probability Estimation\label{sec:cpe}}
In \cite{mohamed2016learning}, two ratio estimation techniques, class probability estimation and ratio matching, are discussed. We briefly show how to use the class probability estimation technique to estimate $\mathnormal{f}$-divergence, and refer readers to the original paper \cite{mohamed2016learning} for the ratio matching technique.

The density ratio can be computed by building a classifier to distinguish between training data and the data generated by the model. This ratio is  $\frac{p(x)}{q_\theta(x)}=\frac{p(x|y=1)}{p(x|y=0)}$, where label $y = 1$ represents samples from $p$ and $y=0$ represents samples from $q$. By using Bayes rule and assuming that we have the same number of samples from both $p$ and $q$, we have $\frac{p(x)}{q_\theta(x)}=\frac{p(x|y=1)}{p(x|y=0)}=\frac{p(y=1|x)p(x)}{p(y=1)}/\frac{p(y=0|x)p(x)}{p(y=0)}=\frac{p(y=1|x)}{p(y=0|x)}$.  We can then set the discriminator output to be  $\mathcal{D}_\phi(x)=p(y =1|x)$, so the ratio can be written as $\frac{p(x)}{q_\theta(x)}=\frac{p(y=1|x)}{1-p(y=1|x)}=\frac{\mathcal{D}_\phi(x)}{1-\mathcal{D}_\phi(x)}$. The generator loss corresponding to an $\f$-divergence can then be designed as $\fdiv{p(x)}{q_\theta(x)} = \int  q_\theta(\v{x})f\br{\frac{p(\v{x})}{q_\theta(\v{x})}} d \v{x}=\int  q_\theta(\v{x}) f(\frac{\mathcal{D}_\phi(x)}{1-\mathcal{D}_\phi(x)})=\mathbb{E}_{q(z)}[f(\frac{\mathcal{D}(\mathcal{G}_\theta(z))}{1-\mathcal{D}(\mathcal{G}_\theta(z))})]$.

\end{document}